# Monitoring Trust in Human-Machine Interactions for Public Sector Applications

**Farhana Faruqe, Ryan Watkins, and Larry Medsker**
Human-Technology Collaboration Lab, The George Washington University, Washington, DC, USA

## Abstract

The work reported here addresses the capacity of psychophysiological sensors and measures using Electroencephalogram (EEG) and Galvanic Skin Response (GSR) to detect levels of trust for humans using AI-supported Human-Machine Interaction (HMI). Improvements to the analysis of EEG/GSR data may create models that perform as well, or better than, traditional tools. A challenge to analyzing the EEG/GSR data is the large amount of training data required due to a large number of variables in the measurements. Researchers have routinely used standard machine-learning classifiers like artificial neural networks (ANN), support vector machines (SVM), and K-nearest neighbors (KNN). Traditionally, these have provided few insights into which features of the EEG/GSR data facilitate the more and least accurate predictions – thus making it harder to improve the HMI and human-machine trust relationship.

A key ingredient to applying trust-sensor research results in practical situations, and monitoring trust in work environments is understanding which key features contribute to trust and then to reduce the amount of data needed for practical applications. We used the Local Interpretable Model-agnostic Explanations (LIME) model as a process to reduce the volume of data required to monitor and enhance trust in HMI systems – a technology that could be valuable for governmental and public sector applications. From customer service in government agencies and community-level non-profit public service organizations to national military and cybersecurity institutions, many public sector organizations are increasingly concerned to have effective and ethical HMI with services that are trustworthy, unbiased, and free of unintended negative consequences.

In many everyday contexts, human trust is a critical factor for effective human-machine interaction. HMI includes, for example, the experiences of a person working in a call center (providing information to the public using an intelligent system) to supplement and verify the facts and advice being given; and the experiences of public bus drivers as they navigate the city with GPS maps and traffic technologies, and now sharing the streets with self-driving vehicles. Explainable AI is used to make the system transparent enough to promote trust in the system. Trust, in the context of this research, is a construct represented by the behavioral reactions of humans in response to machine advice (or outputs), which also includes factors of self-confidence, machine confidence, and machine credibility. Before studying the relationship between explainability and trust; however, the algorithms (e.g., neural network) guiding machine advice must be explainable and maintain their predictability. Understanding which features are contributing to trust would reduce the amount of data needed for practical applications. The use of LIME may be important in that process and help enhance trust in HMI systems. More trustworthy products would be invaluable to governmental and public sector applications, ranging from communication and advising the public to transportation systems and national security.

## Introduction

Trust in human-machine interactions is garnering a lot of attention across societal, governmental, and public sector applications. Human trust is critical to optimizing the potential of these systems; after all, if humans do not trust machine outputs (which is a form of advice to the interacting human in many contexts) then the human is more likely to override the machine and lose the opportunity to benefit from the machine's capabilities. Limited but ongoing research is being done to build a trust sensor model for assessing the levels of trust humans have when collaborating with intelligent machines (i.e., real-time feedback on a human's behavioral response that is in line with the recommendation of the machine).

In this context, trust is defined as a human's reaction to the performance of a machine during human-machine interaction (HMI), without attempting to discern motivation, self-confidence, credibility, and other related factors. This research uses electroencephalogram (EEG) and Galvanic Skin Response (GSR) data (collected in prior research on building a human trust sensor model (Akash et al., 2017) to build an artificial neural network (ANN) model to identify critical features that correlate with trustworthiness. As part of the research, we are investigating capabilities of IBM Watson Studio and OpenScale to analyze and monitor trust, explainability, and transparency.

Improving the interpretability of human-machine interaction models (such as a trust sensor model), while maintaining or improving its predictive accuracy, could have many practical applications. For example, many public sector organizations use call centers internally and as an interface with the public. Typically, call center personnel refer to a computer for informational and decisional support in answering questions, which requires a level of trust that the advice being given out is correct and helpful. Critical examples include suicide hotlines, IRS advisory services, human resource and employment law guidance, and instructions for COVID-19 financial assistance.

In most any of these examples, technologies (such as for minimally invasive EEG/GSR measurements) could be added to low-cost headsets and wristbands used by call center personnel to communicate with the public and monitor computer support systems. The continuous collection of data with trust analysis software could then monitor real-time changes in an employee's trust level and give early-warning feedback to the employee and supervisor when, for instance, experienced employees become uncertain of the accuracy of the information provided by the machine. Post-analysis of daily data collection could allow modification of computer support software and foster ongoing machine learning.

Pervasive, intelligent, and autonomous systems are just starting to reshape our society and the workplace. A study from McKinsey Global Institute indicates that a subset of tasks or activities of almost all occupations globally can be automated (Manyika et al., 2017), increasing the collaborations of people and machines in the workplace. A feasible result is the rise of "human-agent partnerships" (human and intelligent autonomous systems) as a central component of our future society (Jennings et al., 2014). Effective collaboration is required in this scenario, and human trust is a key component for successful human machine interaction (Lee and See, 2004).

## Related Work

### Trust in Human-Machine Interaction (HMI)

Trusting relationships form between humans, and also among humans and machines. Though similarities exist between interpersonal (human-to-human) trust and the trust in automation (human-machine), some important differences have also been documented (Lee See, 2004; Hoff and Bashir, 2015). Trust, self-confidence, machine confidence, and machine credibility are closely related factors (as illustrated by Blasch et al., 2014), and this research focuses specifically on the behavioral component of human reactions to the guidance provided by the machine (i.e., do what the machine suggests, or not; Akash, 2017). This behavioral component of trust provides the most immediate and measurable indicator of trust, an essential feature for the development of an interpretable, real-time trust sensor model.

Measuring human trust of machines *a priori*, by filling out surveys, for instance, is less than optimal for improving HMI since it delays improvements and is subject to a variety of perceptual biases. On-going research, however, attempts to measure real-time trust levels in human-machine interactions (HMI) (Akash et al., 2017, 2108; Hu et al., 2016). For example, Akash (2017) had participants complete a driverless car simulation in which the simulations created patterns of behaviors that should augment the person's trust in automation. EEG was used in the Akash research to capture brain waves of humans while they interacted with a machine, while GSR data was collected to account for anxiety and cognitive load. Using the data captured from EEG and GSR, researchers applied classification algorithms to obtain trust levels (categorical level: trust, distrust) relating the brain activity and skin responses to the responding behaviors of the participants to initiating behaviors of the simulation. This research illustrates the potential for real-time measures of behavior-based trust.

HMI trust is described by Hoff and Bashir (2015) as generally having three common components: trustor, trustee, and the involvement of some sort of task that needs to be performed by the trustee. Hoff and Bashir (2015) also considered the different factors that are capable of influencing human-machine trust and reliance, and their concluding remark is that human-machine interactions are increasing as humans are getting more dependent on automation systems. These interactions are critical to the formation of trust in automation to ensure proper use and minimize machine related accidents (Hoff and Bashir, 2015). This framework of trust can be used in real-life scenarios to build a trust sensor model and to measure trust level in a human when the human is interacting with a machine (in this paper, machine refers to any man made autonomous system). Operationally, trust is reflected in the chosen behavior of a trustor in response to their relationship with the trustee.

### Trust-sensor Model

Hu et al. (2016) describe the design of an empirical trust sensor model as an important first step to measuring trust in HMI. The paper highlights that "the proposed methodology for real-time sensing of human trust level will enable machine algorithm designs aimed at improving interactions between humans and machines" (Hu et al., 2016, p. 2). Hu et al. (2016) demonstrated initial evidence to support that psychophysiological signals (i.e., EEG and GSR) can be used for sensing trust in real-time.

To map human trust level with psychophysiological measurement, Hu et al. first selected the best ten features from a large set of features from EEG data by applying statistical methods. By using various classification models (linear discriminant, linear support vector machine, logistic regression, quadratic discrimination, weighted k nearest neighbors) on the selected best ten features, human trust level was categorized into two groups; trust and distrust. The outcome of the model was promising since it indicated that psychophysiological measurements can be used for sensing real-time trust. However, the selected ten features are likely not suitable for all human subjects since important EEG features can vary substantially across individuals. Given that we live in a multicultural and diverse global society, robust approaches (individualized beyond demographic categorizations) should be sought out so that trust-sensor models can apply to the largest number of potential users possible at this time.

Akash et al. (2017) proposed a model capable of capturing both learned and dispositional trust factors so that it can deal with the dynamic changes of human trust in the con-text of human-machine interactions. In their research, EEG and GSR data were collected while participants (n = 45) reacted to outputs (advice) from autonomous car simulation. Akash et al. (2017) also incorporated demographic data so that the trust sensor model will be suitable for different societies. This is very important since globalization requires deploying autonomous systems to different societies. LTI (linear time-invariant) model is established based on psychological literature and Akash et al. used "gray-box system identification" (p. 1) techniques, which also (in addition to a sample of 45 local participant data) supported a large set (over 500 Amazon Turk participants from the US and India) of human behavioral data to estimate the trust parameter. The trust model successfully provided statistical evidence that there was a difference in trust based on a variety of demographics. Akash (2017) categorized the responses from the participants into two bins corresponding to their demographics: US versus India (national culture) and male versus female (gender). Overall, the Akash (2017) demonstrated that this model can capture the dynamic aspects of human trust more accurately since authors considered expectation bias and cumulative trust. This is an important attribute considering the cumulative effect of participant's previous experience with a machine to assess the present trust level.

## EEG (Electroencephalogram) and GSR (Galvanic Skin Response)

Brainwaves are a continuous output of small electric signals of the brain and its varying frequency and amplitude. In a set period of time, the number of peaks (wave cycle) that occurs in an EEG pattern is called its frequency. Hertz (Hz) is defined as the number of wave cycles that occur in a second and this is useful to determine the normal and abnormal EEG rhythms. There are four frequencies that are generally associated with some psychological and behavioral states:

Delta: 0 – 4 Hz, Theta: 4 – 8 Hz, Alpha: 8 – 16 Hz, Beta: 12 – 30 Hz

EEG signals are non-stationary, noisy, and highly dimensional. All these issues should be considered during data processing, feature extraction and data analysis. Fedjaev (2017), for example, applied deep learning algorithms (a type of ANN) to analyze EEG data in their research and provided good descriptions as to how a deep learning algorithm is a good candidate in modeling the sequential data from EEG signals. ANN is capable to handle noisy data and existing research has supported this point (Soboh et al., 2014; DeGroff et al., 2001).

A GSR signal has two components, the phasic component and the tonic component. The phasic component is accountable for rapid changes in the signal, and the tonic is accountable for the slow changes in the conductance of the skin as influenced by eccrine sweat glands. Typically, GSR is used in recognizing human emotion (Ayata et al., 2016; Yoo et al., 2005).

## Methods

The Method section includes overall strategy, ANN architecture, dataset, data preparation, feature selection, machine learning model building, evaluation, and interpretation.

### Strategy and Design

We used IBM Watson Studio to train the model, StandardScaler, recursive feature elimination (RFE), ANN, K-fold cross validation, and LIME to standardized features, select features, build and evaluate the model and also to explain the predictions.

ANN is a supervised machine learning method consisting of input layer, output layer and one or more hidden layers in between. Neural networks are powerful because of their adaptive learning ability. We have used a multi-layer perceptron (MLP) ANN to perform classification (trust, distrust). Multi-layer feed-forward networks are proven to be universal approximators (Cybenko, 1989; Hornik et al., 1989). Training an ANN, however, routinely requires more efficient and powerful resources than a personal computer can provide, and we therefore used the IBM Watson Studio Lite version. Other algorithms such as recursive feature elimination (RFE), and LIME have been applied to select features, and also to explain the predictions. The explanation step was added to make the model more transparent (i.e., rather than keeping it as a "black box"). Throughout this paper we use the term "feature" to refer to a measurable property in machine learning (input variable). RFE and LIME contributed to feature selection, and this was an iterative process to identify the specific features with an improved performance matrix. To build the Multi-layer Perceptron (MLP) classifier, Python general libraries including scikit-learn library and Keras were used. Following are two architectures of the proposed model; the first one is a shallow network and the second one is a deep network:

- Model 1: Input layer → hidden layer (100 nodes, activation = "relu") → hidden layer (100 nodes, activation = "relu") → output layer. [loss = "binary_crossentropy", optimizer = "rmsprop"].

- Model 2: Input layer → hidden layer (100 neurons, activation = "relu") → hidden layer (100 neurons, activation = "relu") → dropout layer (.2) → hidden layer (100 neurons, activation = "relu") → hidden layer (100 neurons, activation = "relu") → output layer. [loss = "binary_crossentropy", optimizer = "adam"].

### Dataset and Data Processing

To build an HMI trust sensor model (Akash et al., 2017), the research team of the Jain Research Lab at Purdue University collected EEG and GSR data from 48 adult study participants with age ranging from 18 to 46 years. From their dataset, we used the extracted features including EEG and GSR data that we received in csv files. There are two hundred input features, and two outcome classes "trust" (machine's credibility in decision making) and "distrust" (machine's implausibility in decision making) are included in the dataset. The data in the files were organized in two domains: time and frequency. Time-domain and frequency-domain both have these EEG features: mean, variance, peak-to-peak value, mean frequency, root mean square value, and signal energy. However, data from three participants were not usable because of noise. As such, they (and subsequently we) used data from the remaining 45 participants. The data collection experiment, data clean-up and initial feature extraction are described in detail by Aksash et al. (2017).

Table 1: **Sample dataset (out of 200 features)**

|   | x1 | x2 | x3 | x4 | x5 | etc. | y (trust or distrust) |
|---|---|---|---|---|---|---|---|
| 1 | -0.14798 | 15.9287 | 21.1605 | 0.21495 | 0.21495 | 0.58008 | 1 |
| 2 | -0.092273 | 21.8456 | 25.9132 | 0.12129 | -0.1479 | 0.52782 | 1 |
| 3 | -0.3798 | 15.9287 | 21.1605 | -0.0686 | 0.44713 | 0.509 | 1 |
| 4 | -0.065817 | 23.6838 | 25.1141 | 0.1784 | -0.62327 | 0.50485 | 0 |
| 5 | -0.092273 | 15.8456 | 21.1301 | 0.4901 | -0.14798 | 0.52839 | 1 |

For the purpose of this research, the dataset has been split into two groups: 70% for training and 30% for validation. Out of 45 participants, data from 31 individuals were considered for training and the others 14 for validation. One of the recommended data preprocessing steps is data scaling. Scaled data helps to improve performance of the model by avoiding unstable learning from the data (Brownlee, 2018). We have used StandardScalar (Python class) to standardize the features. We also made sure that outcome classes are balanced in both training and validation dataset. At the beginning it was not balanced so we down-sampled it and we have a little over four thousands records to work with. In addition to that, we encode the categorical outcome classes to numbers ("trust"→1, "distrust"→0) before we build and evaluate the the model (Brownlee, 2018). We included a sample dataset in Table 1.

### Feature Selection

We initially used RFE for feature selection. Ding et al. (2015) illustrate that selecting relevant features using RFE can improve the performance of a classification algorithm by avoiding redundant (or not relevant) comparisons of features when using brain signal related data. This RFE algorithm has a hyperparameter to get a number of features and for this scenario it shows that the optimal range was between four to twelve features for this data. One of the RFE objects, "support" identifies which feature is important by adding true or false values to a specific column of the feature. We used this "support" object to select relevant features.

LIME is a powerful algorithm capable of explaining the predictions of any classifier (Ribeiro et al., 2016), and we used LIME to explain our classifier model. This is becoming a popular method (Mittelstadt et al., 2019; Doshi-Velez et al., 2017) to explain the internal decision-making process of a machine learning model and making it transparent to the audience. We have trained RFE and ANN separately with all the features (200). After training, we have selected twelve features from RFE (as mentioned earlier, we got the optimal range for features from RFE). For ANN we have designed and trained both models and we have picked model number two above (deep learning model) for the rest of the experiment because of its better performance. After training the ANN model, we have used LIME with ANN. LIME is a model-agnostic method so this method can be used for any ML (machine learning) model, in this case ANN.

LIME provides the explanation of the predictions of an individual record, one at a time (as illustrated with an example individual record in Figure 1). In the example, LIME explains the prediction of a randomly selected individual record from the validation dataset. The first section includes a prediction probability value for each classification. The predicted probability by LIME for this individual record is 91% in classifying the record to class 1. This prediction is correct (we compared this with the original outcome). Second and third sections provide the list of features with relative weights that are color coded by class (blue is for class 1, orange is for class 0) and provide evidence of the contribution to prediction. Because of this, LIME could also be informative in terms of feature selection.

We ran additional predictions using LIME to create a list of influential features. After that, we compared this list with the selected features from RFE and found some common features in both the lists. With these two lists of features, we have created several combinations of features (e.g., combination_1 with four features, combination_2 with ten features, combination_3 with twelve features) to run the model. We tested out overall ANN performance on the dataset with the different feature combinations as a next step. Based on feature size and performance, we have selected ten features as our final reduced feature list (Table 2). To reduce the number of features, it was essential to begin with an analysis of the 200 original features in order to derive influential features based on comparative accuracy and overall model performance, as described in the results section.

## Results

We trained and validated the deep learning ANN with the selected features (Table 2). To optimally train a machine learning model, it requires tuning the hyperparameter (i.e., choosing the right combination of values for the learning algorithm). This is an iterative and time consuming process, even with IBM Watson Studio. Among all the values tried, this combination (learning_rate=0.01, dropout_rate=0.2, batch_size= 64, epoch= 130, optimizer=adam) provided the best outcome.

We used k-fold cross validation to evaluate the model, where k=10. This validation randomly shuffles the dataset, divides them into 10 groups and applies other steps to esti-mate the strength of the model when it makes predictions on unseen data. As a performance matrix, we have reported the following (including max, mean, min, and SD):

- Accuracy: correct predictions/total population
- F1 score: 2*((precision*recall)/(precision+recall))
- Recall: True positives/(True positives +False negatives)
- Precision: True positives/(True positives + False positive)

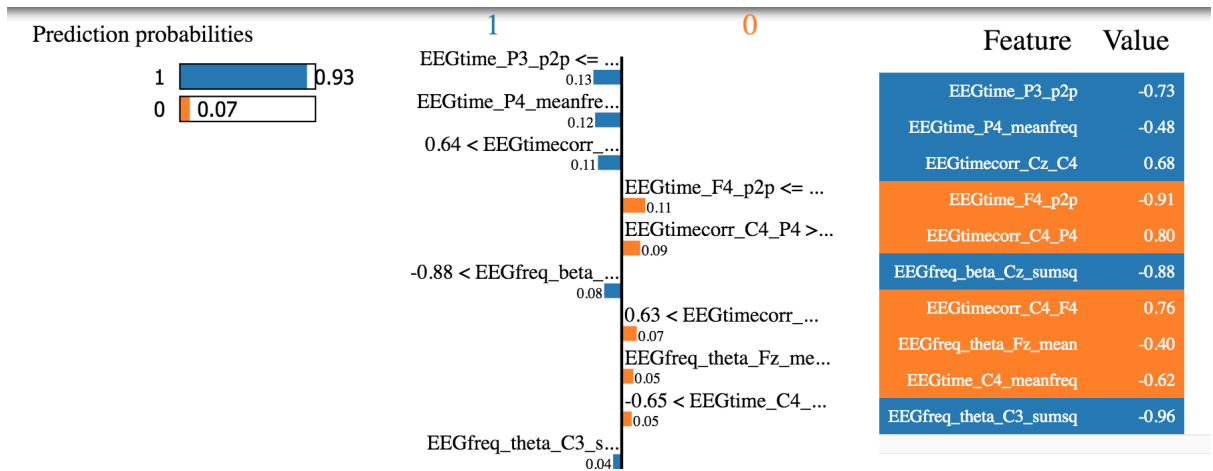

Figure 1: A single record prediction interpretation using LIME

Table 2: **Selected ten features out of two hundred features for the trust-sensor model**

|   | Domain    | Features                  |
|---|-----------|---------------------------|
| 1 | Time      | Peak-to-peak - P3         |
| 2 | Time      | Mean Frequency – P4       |
| 3 | Time      | Mean Frequency – C3       |
| 4 | Time      | Mean Frequency – C4       |
| 5 | Time      | Correlation - C3_C4       |
| 6 | Time      | Correlation - C3_F3       |
| 7 | Time      | Correlation - C4_F4       |
| 8 | Time      | GSR_MaxPhasic             |
| 9 | Frequency | SumSquare of Delta band - POz |
| 10| Frequency | Variance of Beta band - C4 |

Table 3: **Performance matrix of the trust sensor model for validation dataset (10-fold cross validation)**

|      | Accuracy | F1 Score | Recall | Precision |
|------|----------|----------|--------|-----------|
| Max  | 85.16    | 80.56    | 89.06  | 84.91     |
| Mean | 75.00    | 75.45    | 72.02  | 75.18     |
| Min  | 69.53    | 71.21    | 43.75  | 67.16     |
| SD   | 4.4      | 2.64     | 8.47   | 6.12      |

We were able to obtain maximum accuracy of 85% (average accuracy of 75%) using ANN as can be seen in Table 3. Along with the performance, we have identified ten features out of two hundred features for classification.

### Explainability

LIME provides the explanation of the predictions of an individual record, one at a time. The LIME explanation includes prediction probability for classes and a list of features including relative weight. As mentioned in the earlier section, we ran multiple predictions using LIME and created a list of features. The features from the central part of the brain (C3, C4, Cz) were consistent among the test records that we randomly selected. The central part of the brain indicates the emotional aspects of trust behavior. These findings align with other research that has been conducted on the topic of HMI trust (Akash et al., 2017) and mental workload (Dussault et al., 2015) where the role of the central part of the brain is important. GSR measurements contributed one of the features to the analysis, as shown in Table 2, and GSR has been identified as an influential feature to classify trust behavior (Akash et al., 2017). The LIME analysis could, for example, be helpful for a modeller (e.g., end user, subject matter expert, or data scientist) to decide whether to trust the outcome or not, based on greater specification of influential features in the ANN and the individual's prior knowledge or experience in that particular domain. Of course, subject matter experts need to verify the interpretation of the LIME outcome and relate the findings to domain-specific theory. In this way, for instance, explainable machine learning can help promote responsible and ethical AI.

### Discussion

This research potentially makes several contributions in the areas of AI trustworthiness and explainability. The implementation of LIME for identifying specific features of EEG and GSR data associated with trust in HMI shows promise that a reduced number of variables might make applications in HMI more effective and efficient. In this paper we have selected individual records randomly to build the intermediate feature lists for LIME. As a next step, we could select the features from a systematic study guided by an algorithm to explore features from higher and lower accuracy regions. The LIME explanation is helpful for the system designer to understand the reasoning behind decisions and possibly enhance trust in automated systems.

A possible practical outcome, to extend the earlier example, could be the use of lower-cost headsets that include EEG measurements for monitoring trust in HMI communications applications in areas of the public sector like education, health-care, national security, and government agency call centers. Only a small number of selected EEG/GSR signals might be required, reducing the complexity, costs, and intrusiveness of EEG/GSR equipped headsets/wristbands. Explainable AI could have many other practical applications in public and private sector institutions.

## Acknowledgements

We appreciate discussions with Drs. Neera Jain and Kuman Akash at the Purdue University Jain Research Lab and the use of their EEG datasets.